\documentclass[12pt,a4paper]{cibb}

\makeatletter
\providecommand{\@ordinalM}[2]{#1}
\makeatother

\usepackage{subfigure,graphicx}
\usepackage{amsmath,amsfonts,latexsym,amssymb,euscript,xr}
\usepackage{booktabs}
\usepackage[nodayofweek]{datetime}
\usepackage{hyperref}
\usepackage{xurl}
\usepackage{fmtcount}
\usepackage[english]{datenumber}
\usepackage[absolute]{textpos}

\usepackage[table]{xcolor}
\usepackage{color,colortbl,tabularx}

\usepackage[english]{babel}
\usepackage[protrusion=true,expansion=true]{microtype}
\usepackage{amsmath,amsfonts,amsthm}
\usepackage{pifont}

\usepackage{verbatim}
\usepackage{multirow}
\usepackage{float}

\definecolor{LightBlue}{rgb}{0.88,0.9,0.9}

\title{\Large $\ $\\ \bf Thyroid Nodule Segmentation in Ultrasound Images with Conformal Uncertainty Estimation}

\author{\large Francesco Teodori$^{*, 1}$, Pierpaolo Brutti$^{1}$, and Umberto Ferraro Petrillo$^{1}$}
\address{\footnotesize $\ $\\$^1$ Department of Statistical Sciences, Sapienza University of Rome, Rome, Italy. \\
\bigskip
ORCID codes: FT 0009-0005-6284-1061; PB 0000-0003-0458-7868; UFP 0000-0002-4308-5126.
\bigskip
\newline
$^*$corresponding author: francesco.teodori@uniroma1.it
}

\abstract{\small Thyroid Nodule Segmentation, Ultrasound, DeepLabV3+, EfficientNet-B5, Conformal Prediction. 
\normalsize
\\[17pt]
{\bf Abstract} Thyroid nodules affect a large proportion of the general population. 
Although most are benign, a non-negligible fraction is malignant. 
Ultrasound imaging is the primary diagnostic modality, and accurate nodule segmentation is a prerequisite for CAD systems to classify malignancy and assess metastatic risk.\\
In this work, we propose a thyroid nodule segmentation framework that integrates a DeepLabV3+, channel-attention modules and post-hoc conformal prediction, providing both accurate mask and pixel-level uncertainty estimates.
We address thyroid ultrasound nodule segmentation as the first stage of a pipeline that also includes malignancy classification and metastatic risk prediction.
We adopt DeepLabV3+, whose Atrous Spatial Pyramid Pooling (ASPP) captures multi-scale context, with an EfficientNet-B5 backbone for its accuracy-efficiency trade-off. 
Additionally, we test three channel-attention modules: Squeeze-and-Excitation, CBAM and Grouped Channel Attention (GCA), paired with two loss formulations (Dice-BCE and Boundary).\\
Since binary masks do not indicate prediction confidence, we wrap the trained model with a split conformal prediction layer that produces pixel-level uncertainty maps with statistical coverage guarantees.\\
We evaluate all configurations on the TN3K public benchmark using a fixed split and directly compare them with state-of-the-art methods. 
The plain backbone already improves upon previously reported results, with CBAM + Dice-BCE emerging as the best overall configuration.
}

\begin{document}

\renewcommand{\thefootnote}{}
\footnotetext{\small{Article version: \datedate $\;$ h\currenttime  $\;$ CET}}

\thispagestyle{myheadings}
\pagestyle{myheadings}
\markright{\tt Proceedings of CIBB 2026}

\section{Introduction}
Thyroid cancer is the most common endocrine malignancy, accounting for about $3.4\%$ of newly diagnosed cancers worldwide, and approximately $10\%$ of all thyroid nodules are malignant.
The primary diagnostic modality is ultrasound; however, image interpretation involves a subjective component that introduces both inter- and intra-observer variability.
Computer-Aided Diagnosis (CAD) systems can reduce this variability and provide clinicians with a second opinion. 
The first stage of any thyroid cancer CAD pipeline is nodule segmentation: an accurate mask is the starting point for downstream malignancy classification and metastatic-risk quantification.

In this work, we propose a reliable nodule segmentation baseline.
We adopt DeepLabV3+~\cite{deeplab} with an EfficientNet-B5~\cite{efficientnet} backbone pre-trained on ImageNet, and we explore the performance of three channel-attention modules: Squeeze-and-Excitation (SE)~\cite{se}, Convolutional Block Attention Module (CBAM)~\cite{cbam} and Grouped Channel Attention (GCA)~\cite{gcaseg}; combined with two loss functions.
The GCA module has previously been shown to be effective in a multimodal CLIP-based framework for thyroid segmentation~\cite{gcaseg}, but here we evaluate it in a purely visual setting.
Finally, we add a conformal prediction layer~\cite{cp} to the trained model that produces pixel-level uncertainty maps with coverage guarantees.

\section{Method}
\subsection{Datasets and Preprocessing}
We use TN3K~\cite{tn3k}, a well-known open-access dataset of $3493$ thyroid ultrasound images with pixel-level nodule masks, collected from multiple medical centers and devices.
TN3K provides a fixed train-test split, which enables a direct comparison with state-of-the-art results~\cite{prochazka}; we further divide its training set into training and validation ($85\%/15\%$).

Training images are resized to $256\times 256$ and augmented with the Albumentations library: horizontal flipping ($p=0.5$); random brightness and contrast adjustment, both up to $\pm 20\%$ ($p=0.5$); shift-scale-rotation with limits $5\%$, $10\%$ and $20^\circ$ respectively ($p=0.5$).
All images are normalized using ImageNet mean and standard deviation. Validation and test images are only resized and normalized.

\subsection{Architecture}
The segmentation backbone is DeepLabV3+~\cite{deeplab}, an encoder-decoder architecture in which the DeepLabV3 encoder uses Atrous Spatial Pyramid Pooling (ASPP) to capture multi-scale features, while the decoder recovers spatial information and refines object boundaries.\\
The default ResNet-101 encoder is replaced with an EfficientNet-B5 backbone pre-trained on ImageNet.
The EfficientNet family~\cite{efficientnet} jointly scales depth, width and input resolution using a single coefficient $\phi$, while keeping the architecture design fixed; B5 offers a favorable accuracy-to-parameter ratio with respect to the rest of the family.
Additionally, this combination has already been proven effective in~\cite{deeplab+efficientnet}.

\subsection{Additional Modules}
We compare three channel-attention modules placed on top of the backbone: SE, CBAM and GCA. 
EfficientNet-B5 already contains SE blocks inside its building blocks.

\vspace{-0.5em}
\paragraph{Squeeze-and-Excitation (SE)~\cite{se}}
Given an input $\mathbf{X} \in \mathbb{R}^{B \times C \times H \times W}$, SE computes a per-channel descriptor $z \in \mathbb{R}^{C}$ via global average pooling $z_c = \frac{1}{HW}\sum_{i,j} \mathbf{X}(c,i,j)$ and a per-channel weight
\begin{equation}
s = \sigma\big(W_2 \, \mathrm{ReLU}(W_1 z)\big), 
\quad W_1 \in \mathbb{R}^{(C/R) \times C}, \; 
W_2 \in \mathbb{R}^{C \times (C/R)},
\end{equation}
with reduction ratio $R$, producing recalibrated features $\hat{\mathbf{X}} = \mathbf{X} \odot s$.
We add a single SE block at the end of the decoder, just before the classification head.

\vspace{-0.5em}
\paragraph{Convolutional Block Attention Module (CBAM)~\cite{cbam}}
CBAM extends SE to model spatial information.
Channel attention combines average and max pooling through a shared MLP,
\begin{equation}
M_c(\mathbf{X}) = \sigma\big(W_2\,\mathrm{ReLU}(W_1\,\mathrm{AvgPool}(\mathbf{X}))
+ W_2\,\mathrm{ReLU}(W_1\,\mathrm{MaxPool}(\mathbf{X}))\big),
\end{equation}
followed by a spatial map obtained from channel-wise pooling and a $7\times 7$ convolution,
\begin{equation}
M_s(\mathbf{X}') = \sigma\big(f^{7\times 7}\big([\mathrm{AvgPool}(\mathbf{X}'); \mathrm{MaxPool}(\mathbf{X}')]\big)\big).
\end{equation}
The two maps are applied sequentially: $\mathbf{X}' = M_c(\mathbf{X})\odot \mathbf{X}$, $\hat{\mathbf{X}} = M_s(\mathbf{X}')\odot \mathbf{X}'$. 
The spatial component is intended to help localize small nodules.

\vspace{-0.5em}
\paragraph{Grouped Channel Attention (GCA)~\cite{gcaseg}}
GCA partitions $\mathbf{X}$ into $G$ groups along the channel dimension, $\mathbf{X} = [\mathbf{X}_1,\dots,\mathbf{X}_G]$ with $\mathbf{X}_g \in \mathbb{R}^{B\times (C/G)\times H\times W}$ and applies an SE-recalibration independently on each group,
\begin{equation}
s_g = \sigma\big(W_2 \, \mathrm{ReLU}(W_1 z_g)\big), 
\quad W_1 \in \mathbb{R}^{(C/GR) \times (C/G)}, \; 
W_2 \in \mathbb{R}^{(C/G) \times (C/GR)},
\end{equation}
yielding $\hat{\mathbf{X}}_g = \mathbf{X}_g \odot s_g$.
In contrast to~\cite{gcaseg}, where GCA operates on cross-modal features, we evaluate it in a purely visual setting, and we deploy it at three locations in the network: at the encoder bottleneck ($G=16$), on the low-level skip connection feeding the decoder ($G=4$) and at the decoder fusion stage ($G=8$). 
The three locations correspond to high-level context, low-level spatial detail and the fused representation feeding the prediction head.

\subsection{Loss Functions}
The first loss is the standard Binary Cross-Entropy (BCE) loss combined with a Dice term~\cite{bce}.
Given the binary ground-truth mask batch $\mathbf{y}$ and the model logit batch $\mathbf{x}$,
\begin{equation}
\label{BCE}
    \mathcal{L}_{\text{BCE}}(\mathbf{x}, \mathbf{y}) = \mathrm{mean}\{l_1, \dots, l_B\}, \qquad l_i = -\big[\mathbf{y}_i \log\sigma(\mathbf{x}_i) + (1 - \mathbf{y}_i)\log(1 - \sigma(\mathbf{x}_i))\big],
\end{equation}
and the Dice-BCE loss is
\begin{equation}
\label{Dice-BCE}
    \mathcal{L}_{\text{Dice-BCE}} = \mathcal{L}_{\text{BCE}} + \Biggl(1 - \frac{1}{B}\sum_{b=1}^B\frac{\varepsilon + 2 \sum_{h,w} \sigma(x_{b,1,h,w})\, y_{b,1,h,w}}{\varepsilon + \sum_{h,w} \sigma(x_{b,1,h,w}) + y_{b,1,h,w}} \Biggr),
\end{equation}
with smoothing $\varepsilon = 10^{-6}$.

The second loss is the Boundary loss~\cite{boundary}, which penalizes errors more strongly on the nodule frontier, balancing the difference between interior and boundary pixels. 
It is obtained by adding to Eq.~\ref{Dice-BCE} a frontier term $\mathcal{L}_{\text{Frontier}}$ defined as Eq.~\ref{BCE} restricted to boundary pixels (those with a heterogeneous nodule/background neighborhood):
\begin{equation}
    \mathcal{L}_{\text{Boundary}} = \mathcal{L}_{\text{Dice-BCE}} + \lambda \, \mathcal{L}_{\text{Frontier}}, \qquad \lambda \in \mathbb{R}^+.
\end{equation}

\subsection{Training Details}
All models are trained with the same setup: $100$ epochs, batch size $4$, input size $256\times 256$, AdamW optimizer with learning rate $10^{-4}$ and weight decay $10^{-4}$, cosine-annealing schedule with $\eta_{\min}=10^{-6}$ and early stopping with patience $30$ on validation Dice.
The best checkpoint is the one with the highest validation Dice.\\
All experiments were run on a single NVIDIA T4 GPU (Google Colab Pro) using PyTorch and \texttt{segmentation\_models\_pytorch}.

\subsection{Uncertainty Quantification via Conformal Prediction}
A binary mask does not indicate where the model is confident, a relevant limitation since nodule boundaries are intrinsically blurred and subject to inter-observer disagreement.
We therefore wrap the trained network with a split conformal prediction layer~\cite{cp}, applied post-hoc on the predicted pixel probabilities.

We hold out a subset of the TN3K test set as a calibration set \(\mathcal{C}\), disjoint from the images used for the final evaluation.
For each calibration image, we compute a pixel-level non-conformity score that quantifies the disagreement between the predicted probability and the ground truth.\\
Given a target error level $\alpha$, the conformal threshold $\hat{\tau}_\alpha$ is set to the $\lceil(1-\alpha)(|\mathcal{C}|+1)\rceil$-th order statistic of the calibration scores and is used at inference to convert sigmoid probabilities into a prediction set per pixel.
Finally, we compute an inner mask of high-confidence nodule pixels and an outer mask that includes pixels with uncertain classification.\\ 
By construction, the procedure provides a marginal coverage guarantee at level $1-\alpha$. 
The visual interpretation is straightforward: the difference between the outer and inner mask is an uncertainty band concentrated where the model is least confident.

\section{Results}

\subsection{Evaluation Metrics}
We evaluate all models on the TN3K test set ($614$ images) with three families of metrics, all reported in Tab.~\ref{tab:segmentation} and~\ref{tab:detection}.
\begin{table}[H]
\centering
\caption{Pixel-level segmentation metrics on the TN3K test set ($614$ images). $\text{Dice}_g$ is the pixel-level Dice; $\text{mDice}$, $\text{mDice}_{50}$, $\text{mDice}_{10}$ are the per-image mean, median and worst $10\%$. $\text{MCC}_g$ is the global Matthews Correlation Coefficient and $\text{mMCC}$ its per-image mean; both are reported in the $[-1,+1]$ scale. Best per column in \textbf{bold}.}
\label{tab:segmentation}
\setlength{\tabcolsep}{5pt}
\small
\begin{tabular}{c|c|cccc|cc}
\hline
\textbf{\begin{tabular}[c]{@{}c@{}}Attention \\ Mechanism\end{tabular}} &
\textbf{\begin{tabular}[c]{@{}c@{}}Loss \\ Function\end{tabular}} &
\multicolumn{4}{c|}{\textbf{Dice} $\uparrow$} &
\multicolumn{2}{c}{\textbf{MCC} $\uparrow$} \\
\cline{3-8}
 & &
Dice\textsubscript{g} & mDice & mDice\textsubscript{50} & mDice\textsubscript{10} &
MCC\textsubscript{g} & mMCC \\
\hline
\multirow{2}{*}{None}
  & Dice-BCE  & 88.76 & 83.43 & 89.70          & 63.83          & 0.872          & 0.829          \\
  & Boundary  & 89.46 & 84.72 & \textbf{91.19} & \textbf{65.80} & 0.880          & 0.843          \\
\hline
\multirow{2}{*}{GCA}
  & Dice-BCE  & 89.50 & 84.66 & 90.54          & 64.32          & 0.881          & 0.842          \\
  & Boundary  & 89.43 & 84.48 & 90.70          & 64.87          & 0.880          & 0.840          \\
\hline
\multirow{2}{*}{SE}
  & Dice-BCE  & 89.18 & 84.42 & 90.42          & 64.51          & 0.877          & 0.839          \\
  & Boundary  & 88.91 & 83.58 & 89.69          & 61.96          & 0.874          & 0.831          \\
\hline
\multirow{2}{*}{CBAM}
  & Dice-BCE  & \textbf{89.60} & \textbf{84.84} & 90.60 & 65.66 & \textbf{0.882} & \textbf{0.844} \\
  & Boundary  & 89.39          & 84.50          & 90.87 & 64.74 & 0.880          & 0.840          \\
\hline
\end{tabular}
\end{table}
\begin{table}[H]
\centering
\caption{Detection rate (DR) on the TN3K test set ($614$ images). A nodule is detected if at least $50\%$ of its area is covered by the prediction. $\text{DR}_{s/m/l}$ are the detection rates for small, medium and large nodules ($q_{33}=2255$ px, $q_{67}=8062$ px); $\text{DR}_1$, $\text{DR}_M$ are the detection rates for single and multi-nodule images. Best per column in \textbf{bold}.}
\label{tab:detection}
\setlength{\tabcolsep}{6pt}
\small
\begin{tabular}{c|c|ccc|cc}
\hline
\textbf{\begin{tabular}[c]{@{}c@{}}Attention \\ Mechanism\end{tabular}} &
\textbf{\begin{tabular}[c]{@{}c@{}}Loss \\ Function\end{tabular}} &
\multicolumn{3}{c|}{\textbf{DR by size} $\uparrow$} &
\multicolumn{2}{c}{\textbf{DR by count} $\uparrow$} \\
\cline{3-7}
 & &
DR\textsubscript{s} & DR\textsubscript{m} & DR\textsubscript{l} &
DR\textsubscript{1} & DR\textsubscript{M} \\
\hline
\multirow{2}{*}{None}
  & Dice-BCE  & 0.861          & 0.919          & 0.970          & 0.961          & 0.726          \\
  & Boundary  & 0.891          & \textbf{0.946} & 0.980          & \textbf{0.971} & 0.801          \\
\hline
\multirow{2}{*}{GCA}
  & Dice-BCE  & 0.883          & 0.900          & 0.985          & 0.958          & 0.774          \\
  & Boundary  & 0.883          & 0.923          & \textbf{0.990} & \textbf{0.971} & 0.767          \\
\hline
\multirow{2}{*}{SE}
  & Dice-BCE  & \textbf{0.898} & 0.928          & 0.980          & 0.963          & \textbf{0.815} \\
  & Boundary  & 0.876          & 0.928          & 0.975          & 0.967          & 0.753          \\
\hline
\multirow{2}{*}{CBAM}
  & Dice-BCE  & 0.891          & 0.928          & 0.985          & \textbf{0.971} & 0.781          \\
  & Boundary  & 0.865          & 0.905          & 0.985          & 0.960          & 0.740          \\
\hline
\end{tabular}
\end{table}

The first family consists of Dice metrics, computed with different aggregation strategies: $\text{Dice}_g$, the global Dice obtained by pooling pixel-level confusion-matrix entries over the whole test set (biased in favor of larger nodules); $\text{mDice}$, the per-image mean Dice (each image weighs equally), $\text{mDice}_{50}$ and $\text{mDice}_{10}$, the per-image median and worst-$10\%$ percentile, these metrics are reported to show the asymmetric distribution of Dice scores.

The second family consists of the Matthews Correlation Coefficient (MCC). Unlike the Dice coefficient, it also accounts for true negatives (TN), providing a complementary view of model performance.

The third one is the detection rate (DR): a nodule is detected if at least $50\%$ of its area is covered by the prediction. 
We further stratify DR by nodule size: small ($\text{DR}_s$), medium ($\text{DR}_m$) and large ($\text{DR}_l$), with thresholds at the $33$rd and $67$th percentile of the nodule area distribution ($q_{33}=2255$ px, $q_{67}=8062$ px), and by nodule count, distinguishing single-nodule images ($\text{DR}_1$) from multi-nodule images ($\text{DR}_M$).\\
For the sake of brevity, we omit Intersection over Union, Recall and Precision, since they are highly correlated with Dice and would not add information. The same holds for the median and worst-$10\%$ percentile MCC, since they follow the same distribution as the per-image Dice.

\subsection{Models Comparison}

\begin{figure}[H]
    \centering
    \includegraphics[width=0.85\linewidth]{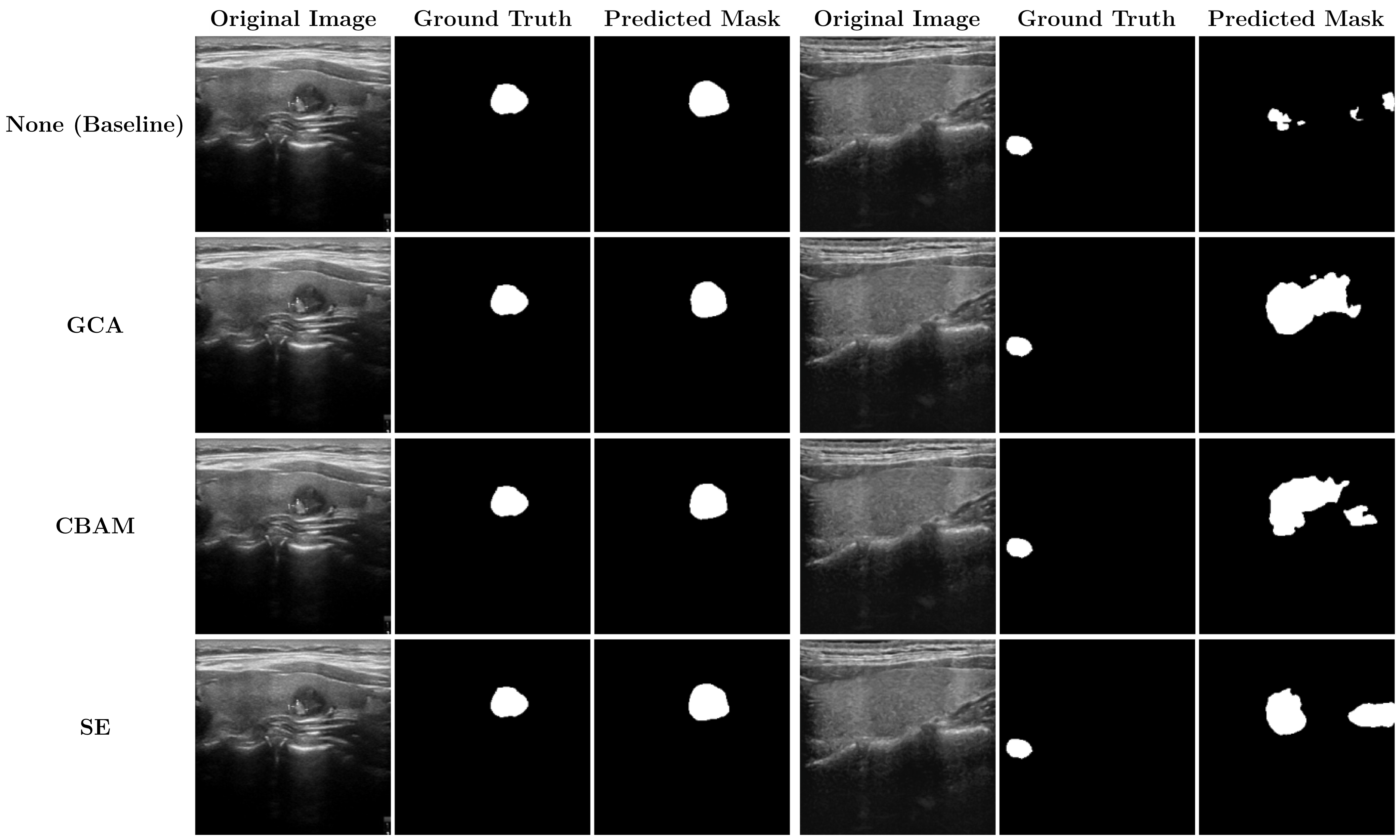}
    \caption{Qualitative segmentations on two TN3K test cases (one easy, one hard) for the four attention configurations, all trained with Dice-BCE. From left to right: original image, ground truth, prediction.}
    \label{fig:qualitative}
\end{figure}
From Tab.~\ref{tab:segmentation},~\ref{tab:detection} and Fig.~\ref{fig:qualitative} we draw four conclusions.
First, several configurations exceed the $83.77\%$ Dice reported
by~\cite{prochazka} on the same test set, with our best mDice reaching
$84.84\%$ (CBAM + Dice-BCE), confirming that the EfficientNet-B5 +
DeepLabV3+ combination is already a strong baseline.\\
Second, no attention mechanism dominates across all metrics. 
Moreover, the per-image median mDice$_{50}$ is approximately $6$ points higher than mDice across all configurations, indicating that the per-image Dice distribution is skewed. The MCC confirms the results given by the Dice coefficients and, as already stated, also the per-image MCC distribution is skewed.\\
Third, detection rates increase monotonically with nodule size ($\text{DR}_s = 0.891$, $\text{DR}_m = 0.928$, $\text{DR}_l = 0.985$ for CBAM + Dice-BCE).
On the other hand, a performance drop occurs in multi-nodule detection rates, since the DRs are $15$--$24$ percentage points lower on multi-nodule images than on single-nodule ones, consistently across all rows.
The difficulty of multi-nodule segmentation is a known issue in the literature, as qualitatively shown in~\cite{prochazka}; some architectures even explicitly restrict their applicability to the single-nodule setting~\cite{deeplab+efficientnet}.\\
Fourth, the qualitative comparison in Fig.~\ref{fig:qualitative} shows that in the easy case, all four configurations recover the mask correctly; in the hard, low-contrast case, all of them fail in similar ways.\\
Considering all metrics jointly, we identify \textbf{CBAM + Dice-BCE} as
the best overall configuration.

\subsection{Uncertainty Visualization}
\begin{figure}[H]
    \centering
    \includegraphics[width=0.85\linewidth]{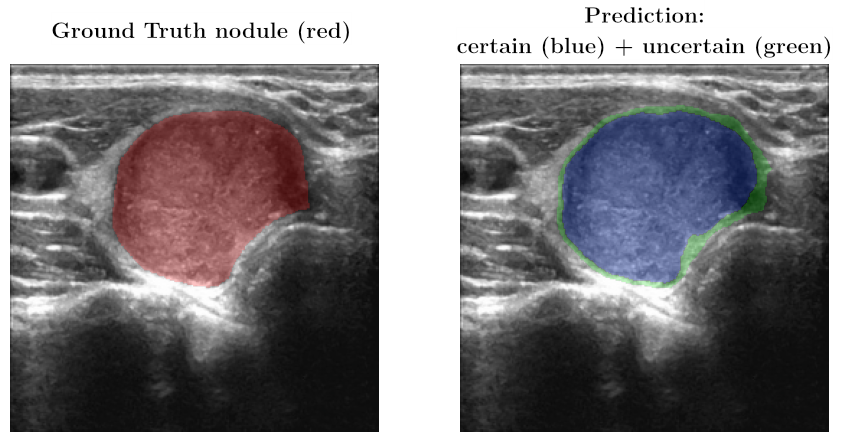}
    \caption{Conformal prediction output on a TN3K test image. Left: ground-truth nodule (red overlay). Right: conformal prediction inner set (blue) and uncertainty band (green) at coverage level $1 - \alpha$.}
    \label{fig:conformal}
\end{figure}

Fig.~\ref{fig:conformal} shows the conformal prediction output (with $99\%$ empirical coverage) on a representative test image for the CBAM + Dice-BCE configuration. 
The blue region is the inner prediction set (high-confidence nodule pixels), while the green band is the difference between the outer and inner set, that is, the pixels with uncertain assignment. 
The uncertainty band concentrates along the predicted boundary, while the interior of the nodule is fully classified as certain. 
This is the desired behavior in a clinical scenario: the model is confident within the nodule and uncertain near the boundary.

\section{Conclusions}
We presented a thyroid nodule segmentation pipeline based on DeepLabV3+ with an EfficientNet-B5 backbone, including an ablation study across three different channel-attention modules (SE, CBAM, GCA), two loss formulations (Dice-BCE and Boundary) and a split conformal prediction layer for pixel-level uncertainty quantification.\\
The plain backbone already matches previously reported results on TN3K, while the additional attention modules provide only marginal, non-monotonic gains: no module dominates across the four Dice variants, the two MCC aggregations and the five detection rates we report.
The conformal layer concentrates the uncertainty along the nodule boundary, aligning the model's confidence with the inter-observer variability intrinsic to the task.
The main limitation is the use of a single dataset.
Future work will extend the pipeline to malignancy classification and metastatic-risk prediction.

\section*{Availability of data and software code}
Our code is available at the following URL: \url{https://github.com/FrankQED/Thyroid-Ultrasound-Nodule-Segmentation-with-Conformal-Uncertainty-Estimation}

The TN3K dataset is available at the following URL: \url{https://github.com/haifangong/TRFE-Net-for-thyroid-nodule-segmentation}
\label{sec:AVAILABILITY}

\footnotesize
\bibliographystyle{unsrt}
\bibliography{bibliography_CIBB_file.bib} 
\normalsize

\end{document}